\documentclass[10pt,twocolumn,letterpaper]{article}

\usepackage{cvpr}
\usepackage{times}
\usepackage{epsfig}
\usepackage{graphicx}
\usepackage{amsmath}
\usepackage{amssymb}
\usepackage{makecell}
\usepackage{enumitem}
\usepackage{hyphenat}

\usepackage[breaklinks=true,bookmarks=false]{hyperref}

\cvprfinalcopy % *** Uncomment this line for the final submission

\def\cvprPaperID{5208} % *** Enter the CVPR Paper ID here

\ifcvprfinal\fi
\begin{document}
\hyphenpenalty=500
%%%%%%%%% TITLE
\title{Assisted Excitation of Activations:\\ A Learning Technique to Improve Object Detectors}
\vspace{4mm}
\author{
\parbox{\linewidth}{\centering
Mohammad Mahdi Derakhshani$^{1*}$,
Saeed Masoudnia$^{1}$\thanks{equally contributed} , 
Amir Hossein Shaker$^{1}$,
Omid Mersa$^{1}$, 
Mohammad Amin Sadeghi$^{1}$, 
Mohammad Rastegari$^{2}$,
Babak N. Araabi$^{1}$
\\
\vspace{4mm}
{\small
$^{1}$MLCM Lab, Department of Electrical and Computer Engineering, University of Tehran, Tehran, Iran.\\
$^{2}$Allen Institute for Artificial Intelligence (AI2)\\
Email: 
mderakhshani, masoudnia, ah.shaaker, o.mersa, asadeghi, araabi\{@ut.ac.ir\}, mohammadr@allenai.org
}
}
}
\maketitle
%\thispagestyle{empty}
%%%%%%%%% ABSTRACT
\begin{abstract}
\begin{figure}[t]
\begin{center}
   \includegraphics[width=8cm]{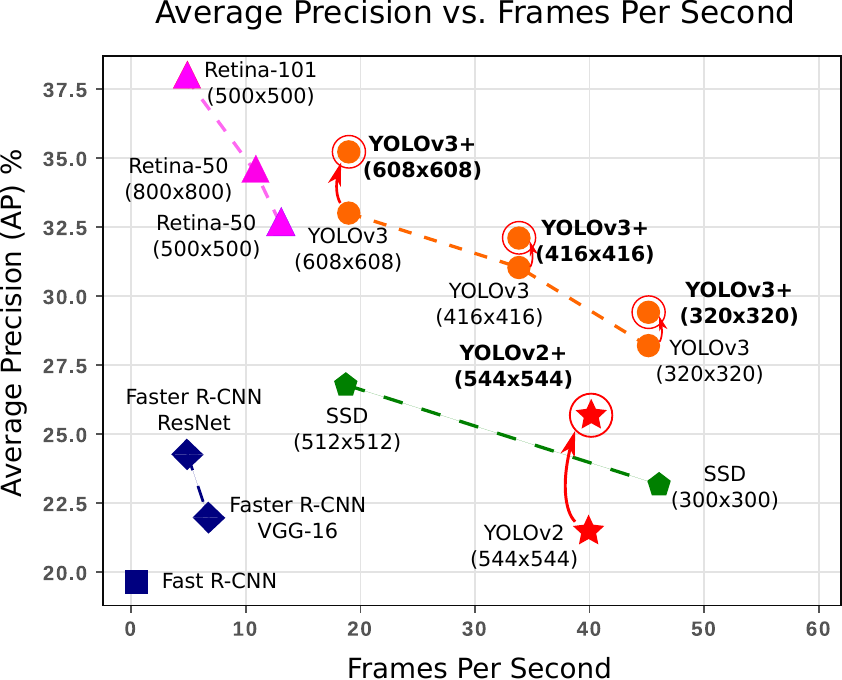}
\end{center}
   \caption{Comparison of different object detection algorithms according to their mean Average Precision and speed~(Frames Per Second). Our improvements (YOLOv2+ and YOLOv3+, highlighted using circles and bold face type) outperform original YOLOv2 and YOLOv3 in terms of accuracy. In terms of speed, our technique is identical to YOLOv2 and YOLOv3. We have evaluated YOLOv3+ on three different image resolutions.}
%\label{fig:long}
\label{fig:splash_comparison}
\end{figure}
\begin{figure}[t]
%\vspace{3mm}
\begin{center}
   \includegraphics[width=\linewidth]{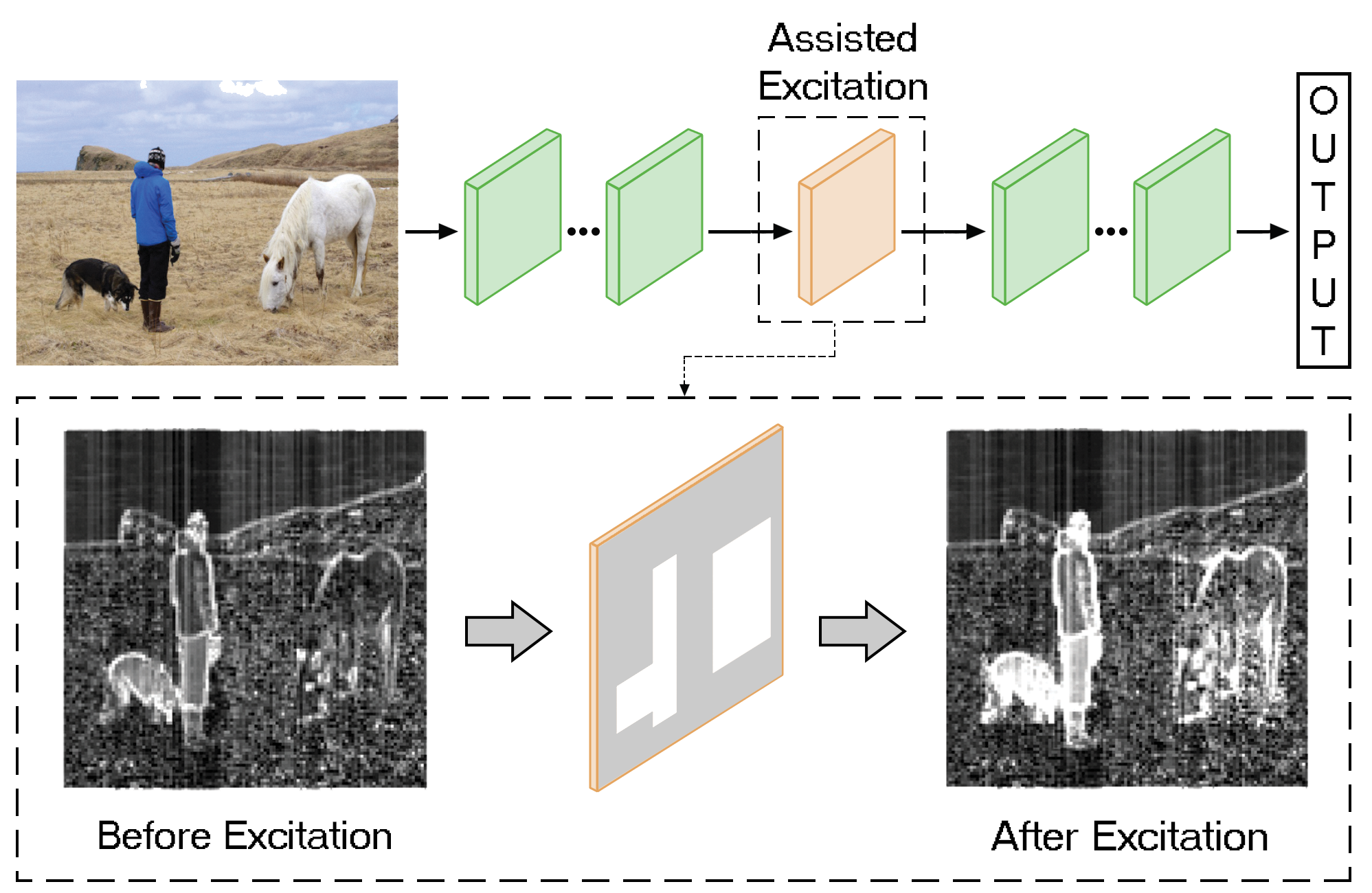}
\end{center}
%\vspace{1mm}
   \caption{An illustration of our proposed Assisted Excitation Module. We manually excite certain activations during training. These activations help improve localization. We excite activations based on object locations. We applied our technique to YOLO object detectors.}
\label{fig:sketch}
\end{figure}
%\vspace*{-2mm}
We present a simple and effective learning technique that significantly improves mAP of YOLO object detectors without compromising their speed. During network training, we carefully feed in localization information. We excite certain activations in order to help the network learn to better localize~(Figure~\ref{fig:sketch}). In the later stages of training, we gradually reduce our assisted excitation to zero. We reached a new state-of-the-art in the speed-accuracy trade-off ~(Figure~\ref{fig:splash_comparison}).\\ Our technique improves the mAP of YOLOv2 by 3.8\% and mAP of YOLOv3 by 2.2\% on MSCOCO dataset.This technique is inspired from curriculum learning. It is simple and effective and it is applicable to most single-stage object detectors.
\end{abstract}
%%%%%%%%% BODY TEXT
\begin{table*}[t]
\centering
\caption{Comparison of the architectures and the characteristics of the three versions of YOLO object detector.}
\vspace{1mm}
\resizebox{\textwidth}{!}
{
\begin{tabular}{|c|c|c|c|c|}
\hline
Model      & Backbone    & Structure         &        Detection Resolution       &   \makecell{Detections\\ Per Grid} 
\\ \hline
YOLOv1 & \makecell{Darknet inspired by GoogleNet~\cite{googlenet}\\ (without inception module) and NIN~\cite{nin}} &  \makecell{24 convolutional layers followed \\ by 2 fully connected layers}            &        Grid of $7 \times 7$    &    2     
\\ \hline
YOLOv2 & Darknet19 inspired by VGG~\cite{vgg} and NIN~\cite{nin} & \makecell{FCN~\cite{fcn} with 19 convolution \\ layers and 5 max-pooling}       &        Grid with stride=$32$      &    5        
\\ \hline
YOLOv3   & Darknet53 inspired by ResNet~\cite{resnet} and FPN~\cite{FPN}           & \makecell{FPN with 75 convolutional layers \\without max-pooling}       &        \makecell{Grids with strides of $32$, $16$ and $8$}      &    3 
\\ \hline
\end{tabular}
}
\label{fig:yolo_characteritics}
\end{table*}
\section{Introduction}
Modern object detectors use Convolutional Neural Networks~\cite{focal,yolo3,faster}. Most of modern object detectors fall into one of two categories: Single-stage detectors (YOLO~\cite{yolo1,yolo2,yolo3}, SSD~\cite{ssd} and Retina-Net~\cite{focal}) and two-stage detectors (R-CNN~\cite{rcnn} and variants~\cite{fast, faster}). Two-stage detectors first generate a number of proposals and then classify them. In contrast, single-stage detectors perform detection in one pass, straight from raw images to final detections. Figure~\ref{fig:splash_comparison} compares a number of notable object detectors according to speed and accuracy. 
YOLO~(You Only Look Once)~\cite{yolo1} is one of the most successful object detector families. These detectors are developed by Redmon et al.~\cite{yolo1,yolo2,yolo3} in three versions: YOLOv1~(2016)~\cite{yolo1}, YOLOv2~(2017)~\cite{yolo2}, and YOLOv3~(2018)~\cite{yolo3}. YOLO detectors are fast and accurate at the same time. They work in real-time and produce high-accuracy detections~\cite{review}. 

While YOLO detectors are very successful, they face two challenges: 1- difficulty in localization~\cite{yolo1,yolo2,yolo3}, and 2- foreground-background class imbalance at training~\cite{focal}. All versions of YOLO face these challenges. In the latest work, Redmon et al.~\cite{yolo3} reported: ``\textit{The performance drops significantly as the IOU threshold increases, indicating YOLOv3 struggles to get the boxes perfectly aligned with the object.}''

Localization problem occurs because YOLO performs classification and localization simultaneously. The last convolutional layer is typically rich in terms of semantics. This is ideal for classification; however, the last convolutional layer is often spatially course for localization. Thus compared to other successful object detectors, YOLO makes more localization errors.

Unlike two-stage detectors, single-stage detectors do not reduce search space to a limited number of candidate proposals. Instead, their search space includes a large number of possible bounding-boxes~(around $10^4$ to $10^5$). Most of these bounding-boxes are negative examples and most of negative examples are easy to classify. As a result, a detector's loss is overwhelmed with easy negative examples while being trained.

This problem was described by Lin et al.~\cite{focal} as foreground-background class imbalance problem. They offered ``focal loss'' to dynamically focus on more difficult negative examples. This loss function greatly improved detection accuracy and resulted in a new model named RetinaNet. Redmon et al.~\cite{yolo3} examined focal loss for YOLOv3, however, they reported that focal loss has been unable to improve YOLOv3.
\newline
\newline
\subsection{Overview of our Solution}
\label{sec:overview}

We propose a solution to address these challenges in YOLO. We only change the way these networks are trained. We propose a technique to \textit{excite} certain activation maps in the network during \textit{training}. We do not change network architecture during inference; we do not change loss function; and we do not manipulate network input or output. 

We test our technique on the training of YOLOv2 and YOLOv3 detectors. During the first epochs of training, we manually excite certain activations in feature maps. Then, in the later epochs of training, we gradually reduce excitation levels to zero. During the last epochs of training, we stop exciting activations. Therefore, the network learns to perform detection without assisted excitation. This strategy is inspired by curriculum learning~\cite{curriculum}; it simplifies the task of detection and localization in the early stages of training and gradually makes the task more difficult and realistic.

We excite activations corresponding to object locations (extracted from ground truth) in feature maps. While we excite these activations, detection becomes easier because our model receives feedback from ground-truth. Therefore, we argue that these excitations help the network 1- improve localization and 2- focus on hard negatives rather than easy negatives. We refer to our method as Assisted Excitation (AE) because we manually excite activations to assist with training.

Our technique helps YOLOv2 improve by $3.8\%$ mAP and YOLOv3 by $2.2\%$ mAP on MSCOCO, without any loss of speed.

%In this article, we first review the related works on YOLO model and past techniques to use auxiliary information in object detectors are first reviewed. The proposed method is then presented in section 3. The experimental results are provided in section 4. Finally, we discuss and conclude the results and propose future research directions in the last section.

%----------------------------------------------------
\section{Related Works}

\textbf{YOLO:} 
Through a sequence of advances, Redmon et al.~\cite{yolo1,yolo2,yolo3} proposed three versions of YOLO. The performance of the latest model is on par with the state-of-the-art. Moreover, YOLO sits at the faster end of the speed-accuracy trade-off. We briefly compare the architecture and characteristics of these versions of YOLO in Table~\ref{fig:yolo_characteritics}.

\begin{figure*}[t]
%\vspace{1cm}
\begin{center}
    \resizebox{\linewidth}{!}{
        \begin{tabular}{cccc}
        \includegraphics[width=0.25\linewidth]{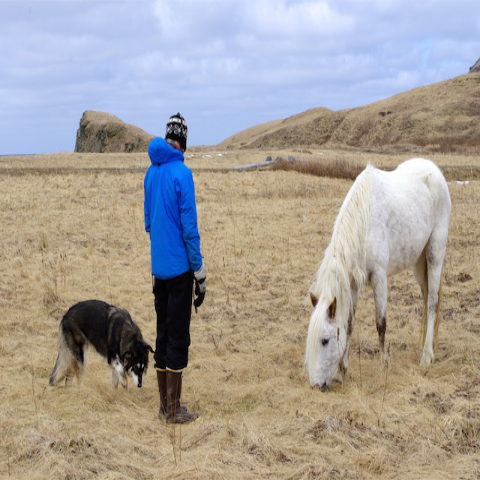}&
        \includegraphics[width=0.25\linewidth]{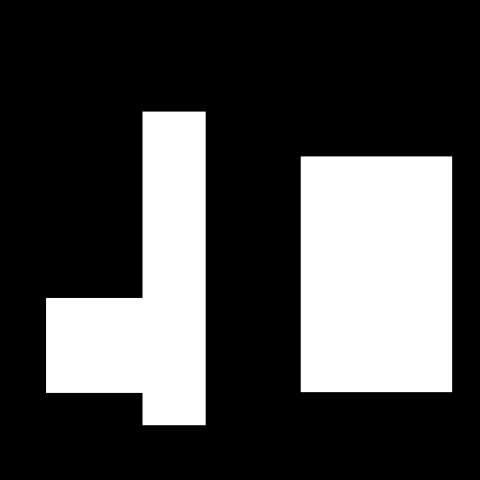}&
        \includegraphics[width=0.25\linewidth]{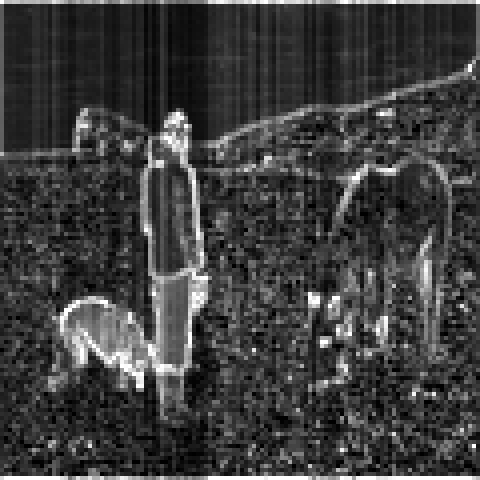}&
        \includegraphics[width=0.25\linewidth]{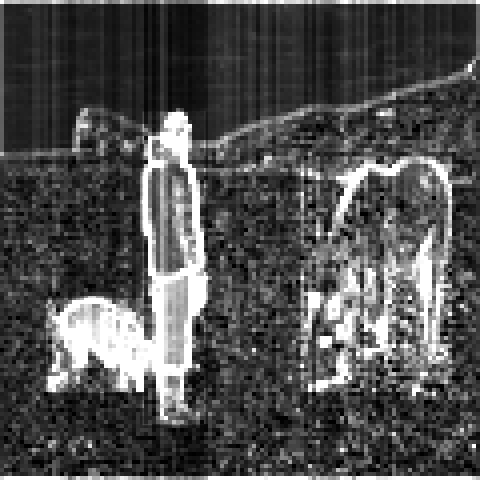}\\
        (a)&(b)&(c)&(d)
        \end{tabular}
    }
\end{center}
   \caption{
        Illustration of our Assisted Excitation process.
        (a) Reference image; 
        (b) Map of object bounding-boxes used to mask excitations;
        (c) Averaged activation before Assisted Excitation Layer;
        (d) Averaged activation after  Assisted Excitation Layer. Please note that excited locations correspond to the object map.}
\label{fig:long}
\label{fig:method}
\end{figure*}

\textbf{Augmenting auxiliary information into CNNs:\newline}
Introducing auxiliary information into CNNs has shown to be useful in certain applications~\cite{context-driven, stackingwithauxiliary, Largeage-gap,CNNforsaliency, text}. A number of works concluded that joint learning of object detection and semantic segmentation can improve both results. These works fall into two categories. The first category~\cite{Simultaneousdetection, Instance-aware} attempts to perform simultaneous detection and segmentation and improve the performance of both tasks~\cite{Simultaneousdetection, Instance-aware, blitznet, multinet, triply}. This combined task is known as instance-aware semantic segmentation.

The second category~\cite{Simultaneousdetection, maskr-cnn, Bottom-upsegmentation, Objectdetectionvia, single-Shot} aims to only boost object detection by introducing segmentation features. Gidaris and Komodakis~\cite{Objectdetectionvia} added semantic segmentation-aware CNN features to detection features at the highest level of R-CNN model. Their model used the auxiliary segmentation information to refine localization. He et al.~\cite{maskr-cnn} proposed Mask R-CNN which extends Faster R-CNN~\cite{faster}.They added a branch for predicting an object segmentation mask in parallel with the existing detection branch. Zhang et al.~\cite{single-Shot} extended an SSD-based object detection model by adding a segmentation branch. However, this branch was trained by weak segmentation ground-truth (box-level segmentation), thus no extra annotation was required.

Several works ~\cite{fuseddnn, illuminating} used the approach of joint segmentation and detection in the application of pedestrian detection. Brazil et al.~\cite{illuminating} also offered multi-task learning on pedestrian detection and semantic segmentation based on the extension of R-CNN. In this model, the weak box-based segmentation mask is infused with both stages of R-CNN model.

Among the reviewed studies, our proposed method is more related to~\cite{illuminating, single-Shot}. Similar to their approaches, we also employ weak segmentation ground-truth only during training and the model efficiency is not affected in our inference phase. Another similarity lies in the fact that there is no need for extra annotation rather than weakly annotated boxes in the detection annotation.

Although the previous studies~\cite{illuminating, single-Shot} developed their models based on R-CNN and SSD respectively, our model is built on top of YOLO model. These studies augmented auxiliary segmentation layers with an extra loss function. Our proposed method does not impose extra computational burden in the training phase. Our main novelty lies in the way of incorporating the ground truth information into the CNN. 
%--------------------------------------------------------------

\section{Challenges in Single-stage Detectors}
\label{sec:challenges}

In Section \ref{sec:overview} we described two challenges that YOLO architecture faces. Here we describe them in more details:
\begin{enumerate}
    \item \textbf{Localization Problem}: For the sake of speed, YOLO performs localization and classification at the same time. Final layers of YOLO architecture produce high-level feature maps. These feature maps are ideal for classification because they are semantic and high-level. However, They are not ideal for localization because they are spatially too course. YOLOv3 tries to address this problem by passing on low-level features (from earlier stages) into localization process. However, Redmon et al. acknowledge that all three versions of YOLO suffer from localization problem.
    \item \textbf{Foreground-Background class imbalance problem}: Two-stage detectors first identify a limited number of object proposals and then classify them. The first stage takes care of most of the localization task. Therefore, the search space in the second stage is limited to a number of proposals that have proper localization.
    
    In contrast, single-stage detectors need to search through a large number possible bounding-boxes ($10^4$ to $10^5$). Many of these bounding-boxes include an object, but most of those containing an object are not localized properly. Therefore, the detector has to search through all of these bounding-boxes and find the single bounding-box that localizes the object the best. this problem is described by Lin et. al.~\cite{focal} They propose a new loss function to address this problem. Redmon et al.~\cite{yolo3}examined focal loss for YOLOv3, however, it did not work out.

\end{enumerate}

\begin{figure}[h]
\begin{center}
% \fbox{\rule{0pt}{2in} \rule{.9\linewidth}{0pt}}
\includegraphics[width=0.85\linewidth]{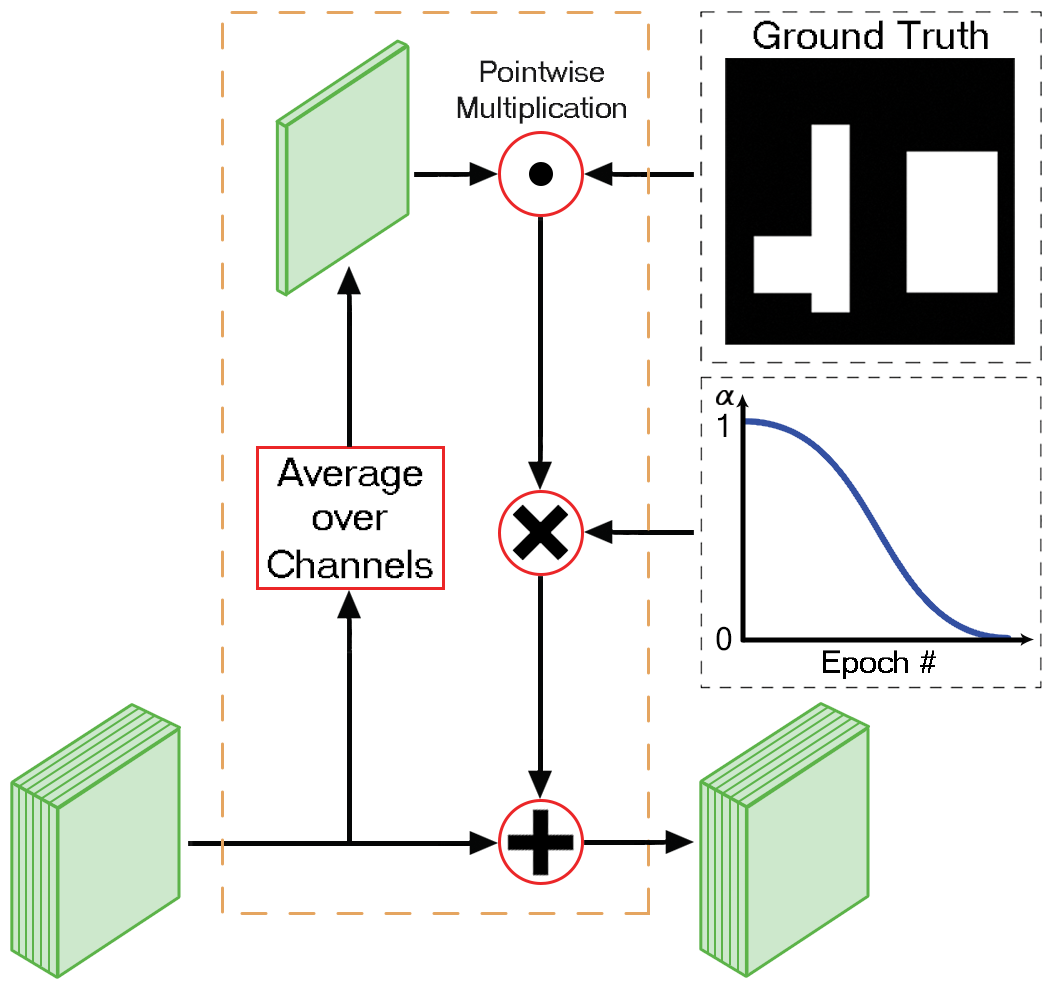}
\end{center}
   \caption{Assisted excitation layer: This layer takes an activation tensor as input. It first averages out all activation maps in input tensor. Then, it masks the results according to object bounding box locations. The excitation value is multiplied by Excitation factor $\alpha$. The result is finally added to each channel of the input tensor and is passed on to the next layer.}
\label{fig:AE}
\end{figure}

\section{Assisted Excitation Process}

We propose a technique to address these challenges. Our technique only applies to the \textit{learning process}. We neither change network architecture nor we change the detection process.

During training, we manually excite certain activations corresponding to object locations. During the initial epochs of training, we perform this additional excitation, however, we gradually decrease the excitation level in the later epochs to zero, see Figure~\ref{fig:AE}.

In the initial epochs of training, our manual activation gives a boost to the best localization bounding-box. This activation helps distinguish the best bounding-box from slightly misplaced bounding-boxes. As we decrease excitation level during the next epochs of training, our model continues to distinguish the best bounding-box from misplaced ones.

We manually excite activations at the locations that we know some object exists. We know where objects exist from the ground-truth annotation, see  Figure~\ref{fig:method}. Ground-truth information is known only during training. Therefore, our final trained model cannot depend on ground-truth. Since we stop manual excitation in the latest stages of training, our model learns to work independent of ground-truth. However, during the initial stages of training, our model depends on a manual excitation that is guided by ground-truth.

These excitations guide the model to 1- improve localization and 2- focus on hard negatives rather than easy negatives. We call our proposed method as Assisted Excitation.

Our technique falls into curriculum learning framework described by Bengio et al.~\cite{curriculum}. The idea behind curriculum learning is that learning space is non-convex, and learning can fall into a bad local-minima. They argue that if we first learn easier tasks and the continue with more complex tasks, we get better performance in terms of the quality of local-minima and generalization.

\begin{figure*}[t]
\begin{center}
% \fbox{\rule{0pt}{2in} \rule{.9\linewidth}{0pt}}
\includegraphics[width=0.8
\linewidth]{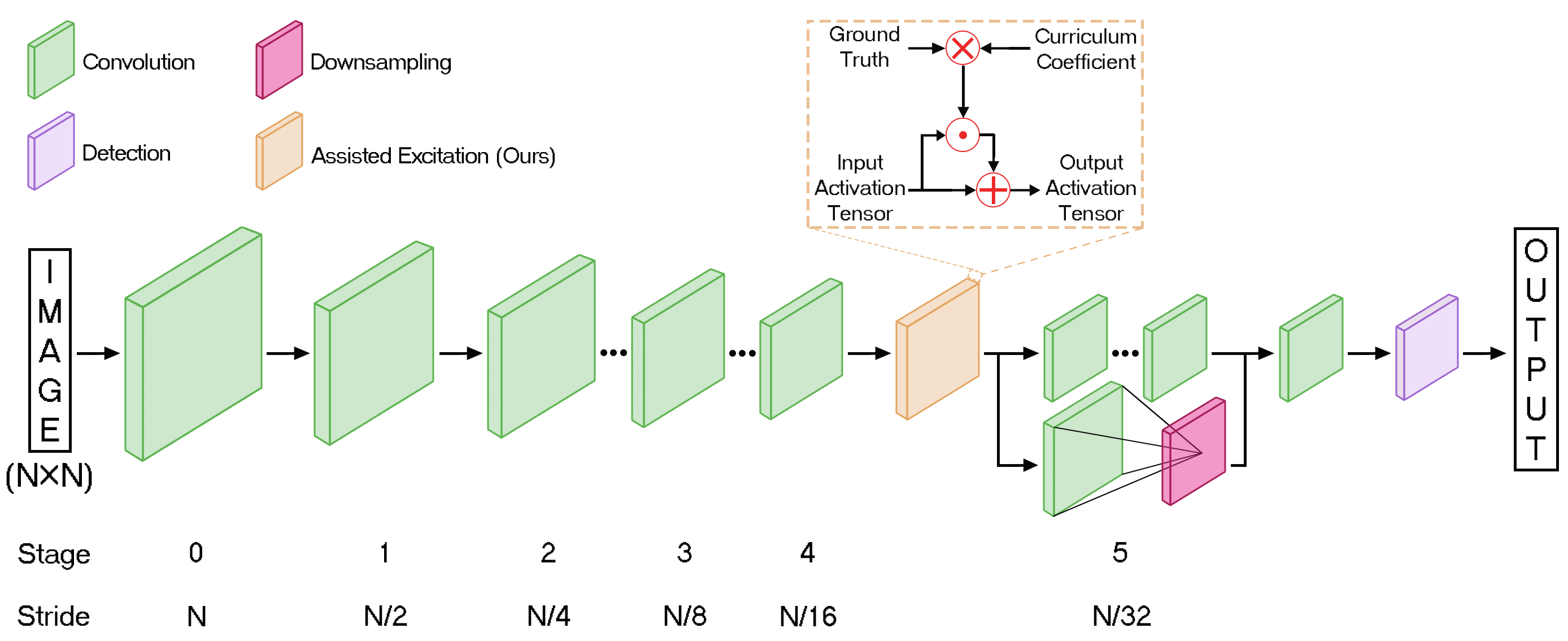}
\end{center}
   \caption{YOLOv2+ architecture. YOLOv2 architecture is modified with our new assisted excitation layer. AE can be added at the end of each stage; Our experiments show that the end of stage 4 is the optimal place for AE. Each stage is composed of a series of activation tensors which have similar resolutions. For example, assume that the input image size is 480x480. Stage 1, stage 2, stage 3, stage 4, stage 5 and stage 6 contain tensors with resolutions 240x240, 120x120, 60x60, 30x30 and 15x15 respectively.}
\label{fig:YOLOv2_plus}
\end{figure*}

\begin{figure*}[t]
\begin{center}
% \fbox{\rule{0pt}{2in} \rule{.9\linewidth}{0pt}}
\includegraphics[width=0.9\linewidth]{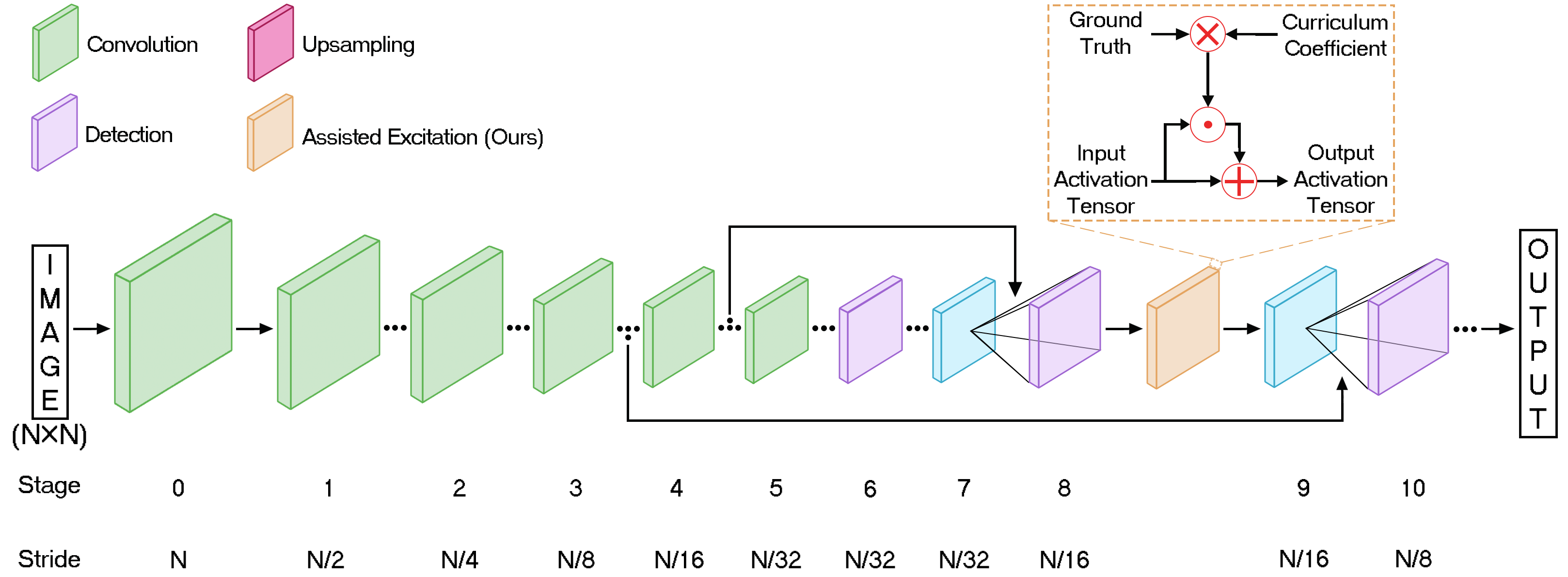}
\end{center}
   \caption{YOLOv3+ architecture. YOLOv3 architecture, which was inspired by~\cite{toward_data_science}, is augmented with an assisted excitation layer. The new layer is added to the end of stage 8.}
\label{fig:YOLOv3_plus}
\end{figure*}

\subsection{Assisted Excitation using Ground-Truth}
\label{sec:AEusingGT}
Assisted excitation can be viewed as a network layer that manipulates neural activations. We can describe an assisted excitation module as follows:
\begin{equation}
\label{eq:1}
a_{(c,i,j)}^{l+1}=a_{(c,i,j)}^{l}+\alpha(t)~e_{(c,i,j)}
\end{equation}
where $a^{l}$ and $a^{l+1}$ are activation tensors at levels $l$ and $l\!+\!1$.\\ $e$ is excitation tensor and $\alpha$ is excitation factor that depends on epoch number $t$. Also $(c,i,j)$ refer to channel number, row and column.
During training, $\alpha(t)$ starts with a non-zero value for initial epochs and gradually decays to zero. $e$ is a function of $a_l$ and ground-truth. To compute $e$, we first construct a bounding-box map $g$ as follows:
\begin{equation}
  g_{(i,j)} =
    \begin{cases}
      1 & \text{If some bbox exists at cell (i,j)}\\
      0 & \text{If ~~no~~\, bbox exists at cell (i,j)}
    \end{cases}       
\end{equation}
% $$
%     g_{(i,j)} = \left\{\begin{array}{ll}
%         1, & ~~~\text{If some bbox exists at cell (i,j)}\\
%         0, & ~~~\text{If ~~no~~\, bbox exists at cell (i,j).}
%         \end{array}\right
% $$
The excitation $e$ in bbox locations can be applied based on different strategies. The straight forward excitation strategy is as follows:
\begin{equation}
\label{eq:2}
e_{(c,i,j)}=\frac{{g_{(i,j)}}}{d}a(c,i,j)
\end{equation}
This strategy excites the activation of bbox location in each channel. Alternative strategy can inhibit out of bbox locations which makes the activations in the bbox locations relatively highlighted.
\begin{equation}
\label{eq:3}
e_{(c,i,j)}=-{(1 - g_{(i,j)})} a_{(c,i,j)}
\end{equation}
These two strategies highlight the activation of bbox locations in each channel independently. We have tried a few variants of this excitation strategy. However, the best performance is not achieved based on these independent manipulation but with the excitation by shared information of bbox locations over all channels. In our method, $e_{(c,i,j)}$ takes an average over all channels of $a_{(c,i,j)}^{l}$. Therefore, it is identical for all values of $c$. We compute excitation tensor $e$ as follows:
\begin{equation}
\label{eq:4}
e_{(c,i,j)}=\frac{g_{(i,j)}}{d}\sum_{c=1}^d a_{(c,i,j)}
\end{equation}
where $d$ refers to the number of feature channels. All the mentioned strategies improve localization. However, the last strategy (Eq~\ref{eq:4}) outperformed the others.
\begin{equation}
\label{eq:5}
\alpha(t)=.5\times\frac{1 + Cos({\pi.t})}{Max\_Iteration}
\end{equation}
Figure~\ref{fig:AE} illustrates our AE layer in more details. 

%Excitation factor $\alpha$ must start from a positive value and gradually decay to zero. There are multiple possibilities for how $\alpha$ decays.
% explain why we chose this alpha configuration

\subsection{Inference}

During inference, $\alpha=0$ and the output of AE layer is identical to its input. Therefore, AE layer is essentially removed during inference. During the final epochs of training, our model learns to function without requiring input from ground-truth. Therefore, we do not use ground-truth information.

In practice, our model architecture is identical to YOLO during inference. Our trained model differs from the standard YOLO model only in model weights. This has two major benefits: \begin{enumerate}\vspace{-1mm}
\item Our trained model is plug and play. We can reuse the heavily optimized detectors developed for all devices.\vspace{-1mm}
\item Our inference time remains identical to the original YOLO detectors while we get better accuracy.
\end{enumerate}

\subsection{Assisted Excitation in YOLOv2 and YOLOv3}

We used Assisted Excitation in YOLOv2 and YOLOv3. For each of the detectors, we performed an ablation study to examine the improvement if we place AE at each stage. We report the results in Experiments section. Figure~\ref{fig:YOLOv2_plus} illustrates the optimal stage for AE in YOLOv2 architecture. Figure~\ref{fig:YOLOv3_plus} illustrates the optimal stage for AE in YOLOv3 architecture.

\begin{figure*}[h]
\begin{center}
   \includegraphics[width=0.95\linewidth]{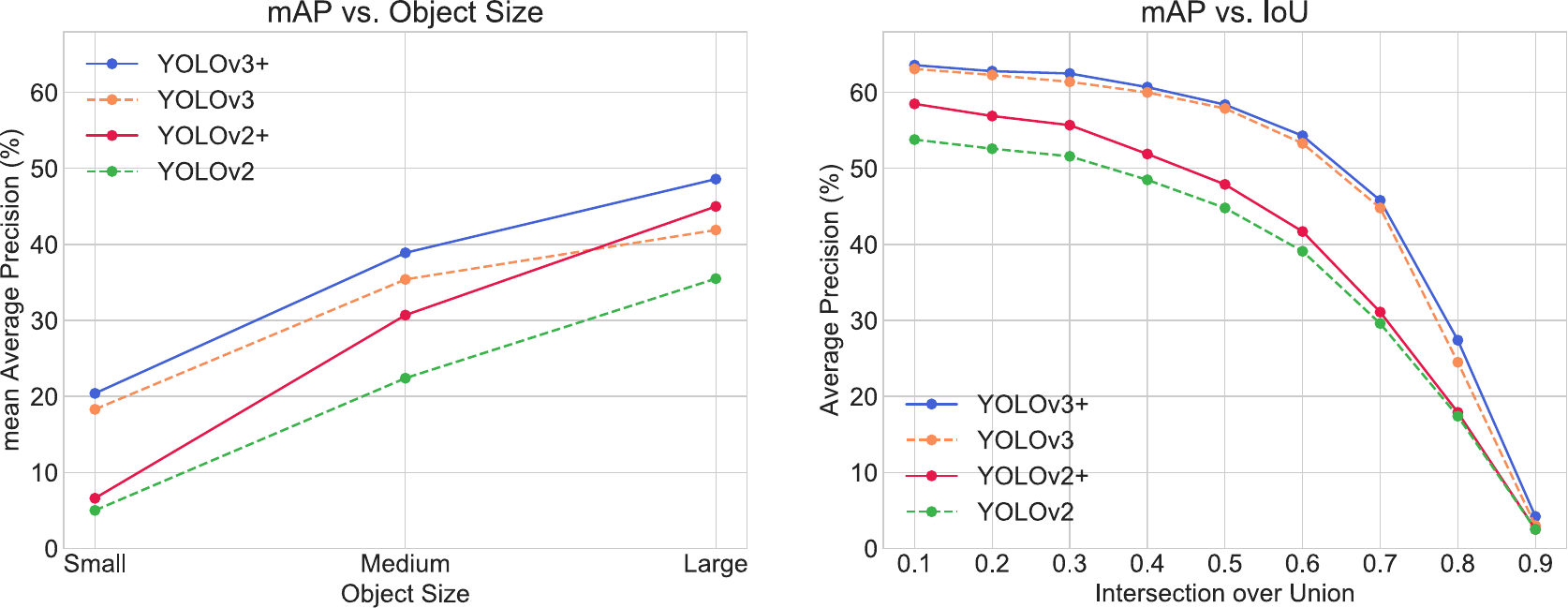}
\end{center}
   \caption{Left: comparison of our proposed methods, YOLOv2+ and YOLOv3+, with their baselines, YOLOv2 and YOLOv3, based on prediction size. As shown, the larger an object is, the more improvement we obtain. Right: comparison of our proposed methods, YOLOv2+ and YOLOv3+, with their baselines, YOLOv2 and YOLOv3, based on Intersection over Union~(IoU) threshold.}
\label{fig:comparisonplot}
\end{figure*}
%-------------------------------------------------------------------------------------------------
\vspace{-2mm}
\section{Experiments and Results}

\textbf{Datasets:} We applied our technique on YOLOv2 and YOLOv3. We evaluated the techniques using two benchmarks: MSCOCO~\cite{MSCOCO} and PASCAL VOC 2007, 2012~\cite{pascalvoc2007}. Similar to the convention of the original YOLO papers~\cite{yolo2,yolo3}, we compare YOLOv2+ with YOLOv2, on PASCAL VOC 2007, 2012 and MSCOCO 2014. Also, we compare YOLOv3+ with YOLOv3 on MSCOCO 2017. Moreover, we also compare with other state-of-the-art detectors on these datasets.

\textbf{Training:} For training, we trained YOLOv2+ and YOLOv3+ from scratch according to the best practices in their original studies~\cite{yolo2,yolo3}. We used Darknet19~\cite{yolo2} and Darknet53~\cite{yolo3} that were pre-trained on IMAGENET dataset, as backbones. Then, we trained whole architectures using Adam~\cite{adam} with initial learning rate of $10^{-5}$, weight decay of $0.0005$, and batch size of $48$. 
%{Refer to YOLO2 YOLO3 implementations on GitHub}

\subsection{YOLOv2+}

% \begin{table}
% \begin{center}
% \caption{The results of applying AGE module through different stages of YOLOv2+. Our proposed model significantly improved the accuracies where applied on the different stages. However, the best accuracy in all terms of AP is achieved in the stage $3$.}
% \resizebox{\linewidth}{!}{
% \begin{tabular}{|l|l|c c c | c c c |}
% \hline
% Method        & Stage       & $AP$       & $AP_{50}$  &  $AP_{75}$    &  $AP_{S}$  &   $AP_{M}$  & $AP_{L}$ \\ \hline
% YOLOv2        & -           & 21.6     & 44.0     &  19.2   & 5.0 & 22.4 & 35.5 \\ \hline
% YOLOv2+ (480) & stage 1     & 24.6	   & 44.8     &	 24.6   & 4.8 & 25.5 & 43.8    \\ \hline
% YOLOv2+ (480) & stage 2     & 25	   & 46  	  &  24.9   & 4.9 & 26.1 & 44.3  \\ \hline
% YOLOv2+ (480) & stage 3     & 25.4	   & 46.9     &	 25.1   & 4.9 & 26.7 & 44.7 \\ \hline
% \end{tabular}
% \label{tab:table2}
% }
% \end{center}
% \end{table}
\begin{table}
\begin{center}
\caption{The results of applying AE module through different stages of YOLOv2+. Our proposed model significantly improved the accuracies where applied on the different stages. However, the best accuracy in all terms of AP is achieved in the stage $4$.}
\vspace{1mm}
\resizebox{0.9\linewidth}{!}{
\begin{tabular}{|l|c|c c c|}
\hline
Method        & Stage       & $AP$     & $AP_{50}$  &  $AP_{75}$\\ \hline
YOLOv2        & -           & 21.6     & 44.0     &  19.2       \\ \hline
YOLOv2+ (480) & stage 2     & 24.6	   & 44.8     &	 24.6       \\ \hline
YOLOv2+ (480) & stage 3     & 25	   & 46  	  &  24.9       \\ \hline
YOLOv2+ (480) & stage 4     & \textbf{25.4}	   & \textbf{46.9}     &	 \textbf{25.1}       \\ \hline
\end{tabular}
\label{tab:table2}
}
\end{center}
\end{table}

\begin{table}
\begin{center}
\caption{The results of different AE strategies in YOLOv2+.}
\vspace{1mm}
\resizebox{1\linewidth}{!}{
\begin{tabular}{|l|c|c c c|}
\hline
Method        & Strategy       & $AP$     & $AP_{50}$  &  $AP_{75}$\\ \hline
YOLOv2        & -   & 21.6     & 44.0     &  19.2       \\ \hline
YOLOv2+ (544) & strategy in Eq.~\ref{eq:2}  & 25.1	   & 45.8     &	 25.8       \\ \hline
YOLOv2+ (544) & strategy in Eq.~\ref{eq:3}  & 24.8	   & 45  	  &  25       \\ \hline
YOLOv2+ (544) & strategy in Eq.~\ref{eq:4}  & \textbf{26} & \textbf{47.9} &	\textbf{25.8}       \\ \hline
\end{tabular}
\label{tab:table3}
}
\end{center}
\end{table}

In order to figure out which layer is the optimal place for our Assisted Excitation module, we performed an ablation study. Table~\ref{tab:table2} lists the accuracy of YOLOv2+ with Assisted Excitation module placed in different stages. 

The best accuracy in all terms of AP is achieved when Advanced Excitation is place in stage $4$. We also examined different excitation strategies discussed in Section~\ref{sec:AEusingGT}. As shown in Table~\ref{tab:table3}, the AE strategy in Eq.~\ref{eq:4} achieved the best result. We will further discuss the results. In the following experiments, we use this configuration (AE on stage $4$) as the default configuration of YOLOv2+.\\
Based on this setting, we  compare YOLOv2+ with YOLOv2 and other current state-of-the-art detectors on MSCOCO test dev-set 2015. The results are compared in Table~\ref{tab:table4}.

\begin{table}
\begin{center}
\caption{The comparison results of YOLOv2+ with YOLOv2 and the other state-of-the-art detectors on MSCOCO test dev-set 2015. The results for the other methods were adapted from. Our proposed YOLOv2+ achieved better accuracies in all terms of APs compared to the previous state-of-the-art detection results.}
\vspace{1mm}
\resizebox{0.95\linewidth}{!}{
\begin{tabular}{|l|l|c c c|}
\hline
Method                                & data & $AP$ &  $AP_{50}$  & $AP_{75}$\\ \hline
Fast RCNN~\cite{fast}                 & train       & 19.7     & 35.9 &    -    \\ \hline
Faster RCNN~\cite{faster}             & trainval    & 24.2     & 45.3 &  23.5   \\ \hline
SSD512~\cite{ssd}                     & trainval35k & 26.8     & 46.5 &  27.8   \\ \hline
YOLOv2~\cite{yolo2} (544)             & trainval35k & 21.6     & 44.0 &  19.2   \\ \hline
YOLOv2+\hfill(480)                    & trainval35k & 25.4     & 46.9 &  25.1   \\ \hline
YOLOv2+\hfill(544)                    & trainval35k & 26	   & 47.9 &  25.8   \\ \hline
YOLOv2+\hfill(608)                    & trainval35k & \textbf{27} & \textbf{50.9} &  \textbf{26}     \\ \hline
\end{tabular}
}
\label{tab:table4}
\end{center}
\end{table}

We compare YOLOv2+ with YOLOv2 using different image resolutions on PASCAL VOC~2007 and VOC~2012. Table~\ref{tab:table4} compares our results with state-of-the-art works on PASCAL. Table~\ref{tab:table5} lists more comprehensive detection results for different resolutions in PASCAL VOC~2007 and 2012.
\vspace{-1mm}
\subsection{YOLOv3+}
Similar to the original YOLOv3 paper~\cite{yolo3}, we conducted several experiments on MSCOCO2017 test-dev dataset. We first report our ablation study on placing Assisted Excitation module on different stages of YOLOv3+.

We compared YOLOv3+ with YOLOv3 in Table~\ref{tab:table6}. As shown in the results, the best performance was achieved where Assisted Excitation module is placed on stage 4. In the remaining experiments, we place AE module on stage 4. Based on this setting, we also compare different image resolutions. Table~\ref{tab:table7} compares YOLOv3+ and YOLOv3 on different input resolutions.Table~\ref{tab:table8} compares our proposed YOLOv3+ with state-of-the-art detectors on MSCOCO2017 test dev-set.

\begin{table}
\begin{center}
\caption{The results for comparison of YOLOv2+ with YOLOv2 in different input resolutions on PASCAL VOC 2007 and 2012. These results were also compared with state-of-the-art detectors on this dataset. Our proposed model significantly improved the accuracy of YOLOv2 in all tested resolutions. YOLOv2+ also 
achieved high accuracy compared to the previous state-of-the-art detection results.}
\vspace{1mm}
\resizebox{0.9\linewidth}{!}{
\begin{tabular}{|l|c|c|c|}
\hline
Detection Frameworks & Train       & mAP      &  FPS        \\ \hline
Fast R-CNN           & 2007+2012   & 70.0     &  44.0       \\ 
Faster R-CNN ResNet  & 2007+2012   & 76.4     &  48.4        \\ 
YOLO                 & 2007+2012   & 63.4     &  26.7         \\ 
SSD500               & 2007+2012   & 76.8     &  26.7          \\ \hline
%YOLOv2 $288 \times 288$   & 2007+2012   & 69.0     &  48.4     \\ 
%YOLOv2 $352 \times 352$   & 2007+2012   & 73.7     &  48.4     \\ 
YOLOv2\hfill(416)  & 2007+2012   & 76.8     &  26.7     \\ 
YOLOv2\hfill(480)  & 2007+2012   & 77.8     &  26.7     \\
YOLOv2\hfill(544)   & 2007+2012   & 78.6     &  26.7     \\ \hline
%YOLOv2+ $288 \times 288$  & 2007+2012   & 74.3     &  48.4     \\ 
%YOLOv2+ $352 \times 352$  & 2007+2012   & 78       &  48.4     \\ 
YOLOv2+\hfill(416)  & 2007+2012   & 80.6     &  26.7     \\ 
YOLOv2+\hfill(480)  & 2007+2012   & 81.7     &  26.7     \\
YOLOv2+\hfill(544)  & 2007+2012   & \textbf{82.6}     &  \textbf{26.7}     \\ \hline
\end{tabular}
}
\label{tab:table5}
\end{center}
\end{table}

\begin{table*}
\begin{center}
\caption{The comparison results of YOLOv2+ with state-of-the-art detectors on PASCAL VOC 2012. The results for the other detectors were adapted from~\cite{yolo2}.  Our proposed YOLOv2+ achieved  better  accuracies  in  all  terms  of  APs  compared  to  the previous state-of-the-art detection results.}
\resizebox{\linewidth}{!}{
\setlength\tabcolsep{3pt}
\begin{tabular}{|l|c|c c c c c c c c c c c c c c c c c c c c|}
\hline
Method               & mAP      &  aero & bike & bird & boat & bottle & bus & car & cat & chair & cow & table & dog & horse & mbike & person & plant & sheep & sofa & train & tv \\ \hline
Fast R-CNN            & 68.4     &  82.3 & 78.4 & 70.8 & 52.3 & 38.7  & 77.8 & 71.6 & 89.3 & 44.2 & 73.0 & 55.0 & 87.5 & 80.5 & 80.8 & 72.0 & 35.1 & 68.3 & 65.7 & 80.4 & 64.2     \\ 
Faster R-CNN          & 70.4     &  84.9 & 79.8 & 74.3 & 53.9 & 49.8  & 77.5 & 75.9 & 88.5 & 45.6 & 77.1 & 55.3 & 86.9 & 81.7 & 80.9 & 79.6 & 40.1 & 72.6 & 60.9 & 81.2 & 61.5     \\ 
YOLO                  & 57.9     &  77.0 & 67.2 & 57.7 & 38.3 & 22.7  & 68.3 & 55.9 & 81.4 & 36.2 & 60.8 & 48.5 & 77.2 & 72.3 & 71.3 & 63.5 & 28.9 & 52.2 & 54.8 & 73.9 & 50.8     \\ 
SSD512                & 74.9     &  87.4 & 82.3 & 75.8 & 59.0 & 52.6  & 81.7 & 81.5 & 90.0 & 55.4 & 79.0 & 59.8 & 88.4 & 84.3 & 84.7 & 83.3 & 50.2 & 78.0 & 66.3 & 86.3 & 72.0     \\ 
YOLOv2 544            & 73.4     &  86.3 & 82.0 & 74.8 & 59.2 & 51.8  & 79.8 & 76.5 & 90.6 & 52.1 & 78.2 & 58.5 & 89.3 & 82.5 & 83.4 & 81.3 & 49.1 & 77.2 & 62.4 & 83.8 & 68.7     \\ \hline
YOLOv2+ 544           &\textbf{75.6}     &  \textbf{87.9} & \textbf{85.1} & \textbf{76.1} & \textbf{62.0} & \textbf{53.7}  & \textbf{81.2} & \textbf{79.2} & \textbf{93.1} & \textbf{53.9} & \textbf{81.1} & \textbf{59.4} & \textbf{90.6} & \textbf{84.7} & \textbf{85.6} & \textbf{84.7} & \textbf{51.4} & \textbf{79.8} & \textbf{64.7} &	\textbf{86.7} & \textbf{71.3} \\ \hline
\end{tabular}
\label{tab:table6}
}
\end{center}
\end{table*}

% \begin{table}
% \begin{center}
% \caption{The results for applying AGE module in different stages of YOLOv3+ on MSCOCO2017 test dev-set. These results were compared with original YOLOv3. Our proposed YOLOv3+ improved the accuracies in all tested stages. However, the best accuracies were achieved for stage 3.}
% \resizebox{\linewidth}{!}{
% \begin{tabular}{|l|l|c c c | c c c|}
% \hline
% Excitation Stage & Stage              & $AP$   &  $AP_{50}$  & $AP_{75}$    &  $AP_{S}$  &   $AP_{M}$  & $AP_{L}$ \\ \hline
% YOLOv3 (608)     & -                  &  33.0  & 57.9        &  34.4        & 18.3       & 35.4        & 41.9     \\ \hline
% YOLOv3+ (608)    & Stage3             &  35.2  & 58.4        &  38.4        & 20.4       & 38.9        & 48.6     \\ \hline
% YOLOv3+ (608)    & Stage4             &  35.1  & 58.2        &  38.4        & 20.3       & 38.8        & 48.4     \\ \hline
% YOLOv3+ (608)    & Stage7             &  34.5  & 58.0        &  37.9        & 18.8       & 38.0        & 47.6     \\ \hline
% YOLOv3+ (608)    & Stage5             &  34.2  & 56.1        &  37.6        & 19.2       & 37.8        & 47.6     \\ \hline
% YOLOv3+ (608)    & Stage9            &  33.5  & 54.6        &  37.1        & 18.6       & 37.1        & 47.1     \\ \hline
% \end{tabular}
% \label{tab:table6}
% }
% \end{center}
% \end{table}
\begin{table}
\begin{center}
\caption{The results for applying AE module in different stages of YOLOv3+ on MSCOCO2017 test dev-set. These results were compared with original YOLOv3. Our proposed YOLOv3+ improved the accuracies in all tested stages. However, the best accuracies were achieved for stage 3.}
\vspace{1mm}
\resizebox{0.85\linewidth}{!}{
\begin{tabular}{|l|c|c c c|}
\hline
Excitation Stage & Stage              & $AP$   &  $AP_{50}$  & $AP_{75}$    \\ \hline
YOLOv3\hfill(608)    &   -                  &  33.0  & 57.9        &  34.4        \\ \hline
YOLOv3+ (608)    & Stage3             & \textbf{35.2} & \textbf{58.4} & \textbf{38.4}        \\ \hline
YOLOv3+ (608)    & Stage4             &  35.1  & 58.2        &  38.4        \\ \hline
YOLOv3+ (608)    & Stage5             &  34.2  & 56.1        &  37.6        \\ \hline
YOLOv3+ (608)    & Stage7             &  34.5  & 58.0        &  37.9        \\ \hline
YOLOv3+ (608)    & Stage9            &  33.5  & 54.6        &  37.1        \\ \hline
\end{tabular}
}
\label{tab:table7}
\end{center}
\end{table}

% \begin{table}
% \begin{center}
% \caption{Ablation study on improvement of YOLOv3+ in different resolution.}
% \resizebox{\linewidth}{!}{
% \begin{tabular}{|l|c c c | c c c|}
% \hline
% Resolution       & $AP$ &  $AP_{50}$  & $AP_{75}$    &  $AP_{S}$  &   $AP_{M}$  & $AP_{L}$ \\ \hline
% YOLOv3  (320)    & 28.2  & 47.7 &  30.0              & 8.8  & 30.9 & 47.4 \\ \hline
% YOLOv3+ (320)    & 29.1  & 50.2 &  30.8              & 9.8  & 32.1 & 48.5 \\ \hline
% YOLOv3  (416)    & 31.0  & 51.0 &  34.1              & 13.0 & 31.1 & 49.3 \\ \hline
% YOLOv3+ (416)    & 32.0  & 53.0 &  34.8              & 13.9 & 35.2 & 49.9 \\ \hline
% YOLOv3  (480)    & 31.6  & 51.2 &  34.5     & 14.0 & 35.0 & 48.7 \\ \hline
% YOLOv3+ (480)    & 32.4  & 53.0 &  35.2     & 14.8 & 36.0 & 49.5 \\ \hline
% YOLOv3  (544)    & 33.1  & 51.8 &  35.9     & 16.3 & 35.5 & 47.5 \\ \hline
% YOLOv3+ (544)    & 33.8  & 55.5 &  37.3     & 18.1 & 37.3 & 49.0 \\ \hline
% \end{tabular}
% \label{tab:table7}
% }
% \end{center}
% \end{table}
%\vspace{-10mm}
\begin{table}
\begin{center}
\caption{Ablation study on improvement of YOLOv3+ in different input resolutions.}
\vspace{1mm}
\resizebox{0.7\linewidth}{!}{
\begin{tabular}{|l|c c c|}
\hline
Method           & $AP$  &  $AP_{50}$  & $AP_{75}$ \\ \hline
YOLOv3\hfill(320)    & 28.2  &    47.7     &   30.0    \\ \hline
YOLOv3+\hfill(320)    & 29.1  &    50.2     &   30.8    \\ \hline
YOLOv3\hfill(416)    & 31.0  &    51.0     &   34.1    \\ \hline
YOLOv3+\hfill(416)    & 32.0  &    53.0     &   34.8    \\ \hline
YOLOv3\hfill(480)    & 31.6  &    51.2     &   34.5    \\ \hline
YOLOv3+\hfill(480)    & 32.4  &    53.0     &  35.2      \\ \hline
YOLOv3\hfill(544)    & 33.1  &    51.8     &  35.9      \\ \hline
YOLOv3+\hfill(544)    & \textbf{33.8}  &    \textbf{55.5}     &  \textbf{37.3}      \\ \hline
\end{tabular}
}
\label{tab:table8}
\end{center}
\end{table}

\begin{figure}[t]
%\vspace{15mm}
\begin{center}
     \resizebox{\linewidth}{!}{
        \begin{tabular}{cc}
        \includegraphics[width=0.30\linewidth]{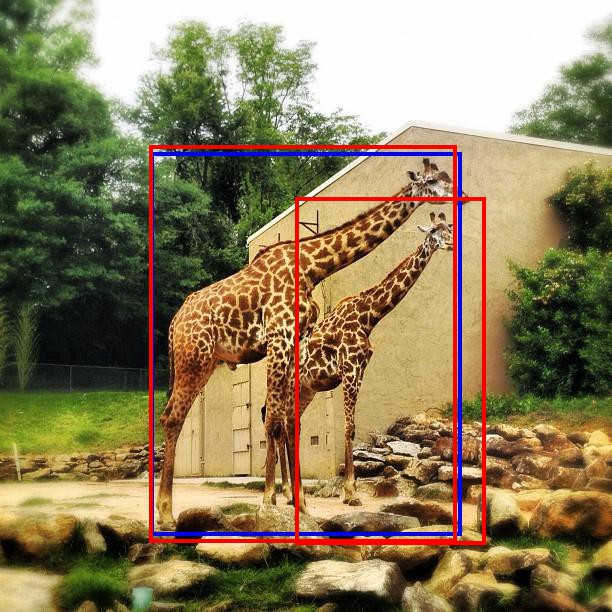}&
        \includegraphics[width=0.25\linewidth]{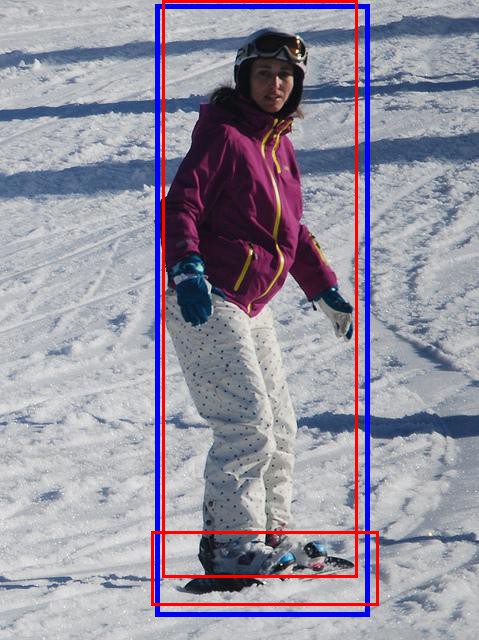}\\
        \includegraphics[width=0.40\linewidth]{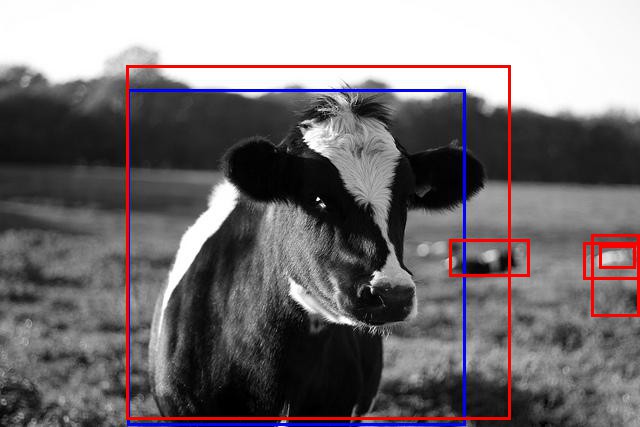}&
        \includegraphics[width=0.35\linewidth]{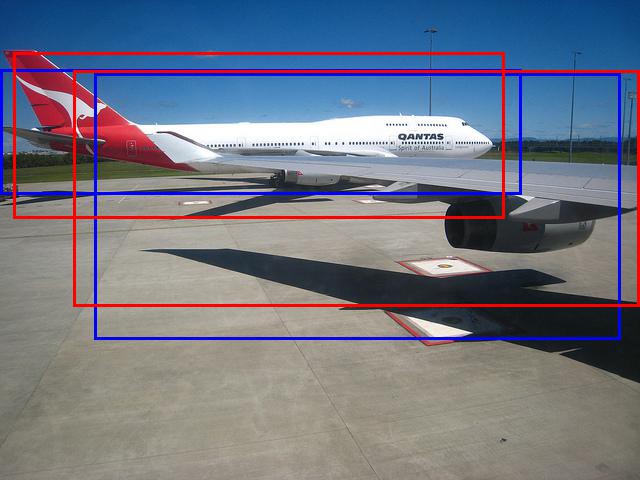}\\
        \end{tabular}
     }
\end{center}
\vspace{-1mm}
   \caption{Visual Comparison of YOLOv2+ and YOLOv2 prediction. As shown, Our proposed method (red bounding box) localizes objects better with respect to YOLOv2's prediction (blue bounding box). In addition to localization, our proposed method increases the number of true positive bounding boxes. }
\label{fig:comparison-yolo2}
\end{figure}
\vspace{-2mm}
\begin{figure}[t]
%\vspace{15mm}
\begin{center}
     \resizebox{\linewidth}{!}{
        \begin{tabular}{cc}
        \includegraphics[width=0.30\linewidth]{images/comp_1.jpg}&
        \includegraphics[width=0.25\linewidth]{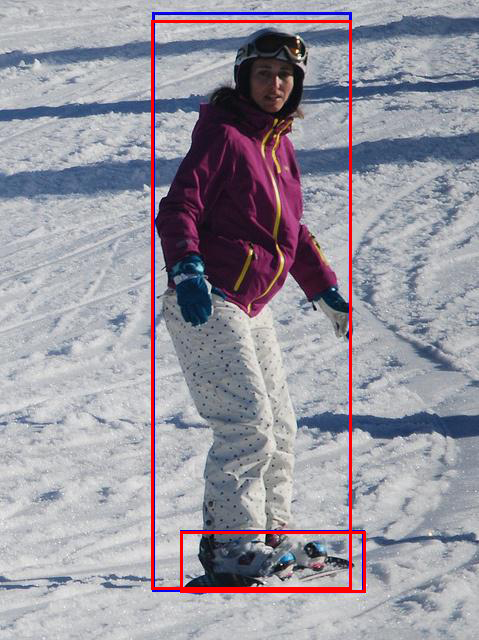}\\
        \includegraphics[width=0.40\linewidth]{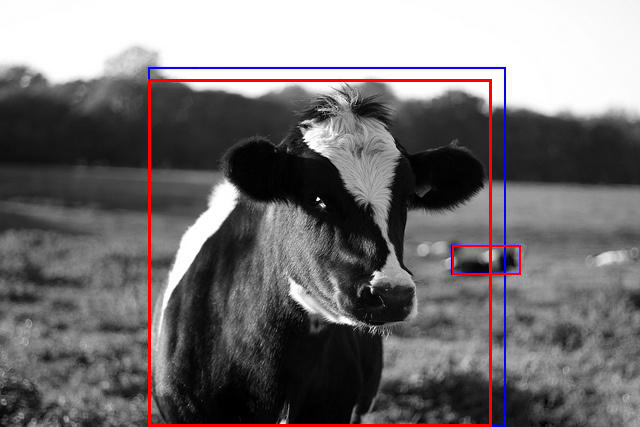}&
        \includegraphics[width=0.35\linewidth]{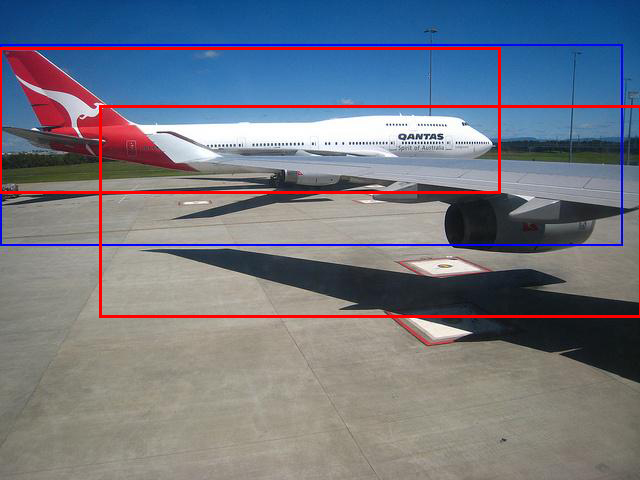}\\
        \end{tabular}
     }
\end{center}
\vspace{-1mm}
   \caption{Visual Comparison of YOLOv3+ and YOLOv3 prediction. As shown, Our proposed method (red bounding box) localizes objects better with respect to YOLOv3's prediction (blue bounding box).}
\label{fig:comparison-yolo3}
\end{figure}
%\vspace{2mm}
\subsection{Localization}

Improvement in localization can be seen in qualitative results.
Figure~\ref{fig:comparison-yolo2} compares localization results between YOLOv2 and YOLOv2+. Figure~\ref{fig:comparison-yolo3} compares localization results between YOLO32 and YOLOv3+. 

Figure~\ref{fig:comparisonplot} right, compares YOLOv2, YOLOv3, YOLOv2+ and YOLOv3+ in terms of mAP versus intersection over union threshold. Figure~\ref{fig:comparisonplot} left, show that our improvement rates increase as the objects become larger. These results in addition to the mentioned theoretical analysis implies that the proposed technique improves the localization ability of YOLO, in specific on medium and large objects.

Our experimental results show that the AE technique improved the accuracy regardless of what stage it is placed at. Further, our experiments show that best improvements are achieved were AE is placed in the mid-level stages. 

In YOLOv2+, the best performance was achieved by placing AE in stage 4 (stride=16). This stage is located in the mid-layers of the model including both localization information and semantic information. In YOLOv3+, the best performance was achieved by placing AE in stage 3 (stride=8). This stage is also located in the mid-layers of the model. The excitation in this stage affects not only in the first detection head but also in both second and third heads because of the skip connections. 

\section{Discussions}
%Here we discuss a few topics regarding our technique.

%The proposed method improves representation learning because:\newline
%- Regarding classification information, the proposed method accumulates information of whole feature maps in one stage along with noise reduction through averaging.\newline
%- Regarding localization information, the activations corresponding to object bounding box locations are amplified compared to remaining areas. It assists forward better discrimination of object locations versus background.\newline
%Nevertheless, excitation of activation maps does not interfere in routine function of the network since mini-batch normalization is used after each stage in YOLO. Therefore, it normalizes the activation rates through each channel.\newline\
\noindent\textbf{Excite object regions vs suppress non-object regions?\newline}
%We discussed foreground-background class imbalance problem in Section~\ref{sec:challenges}. According to this problem, the bulk of our search space consists of negative examples. If we suppress non-object regions, we will affect a large fraction of search space. After we reduce suppression factor to zero at the end of training, the network will need to re-score most of the candidates in search space. In contrast, when we only excite object regions, the network will only need to keep track of much fewer positive examples. Therefore, the model can more easily handle such a change.
We discussed foreground-background class imbalance problem in Section~\ref{sec:challenges}. According to this problem, bulk of our search space consists of negative examples. We proposed different object excitations vs non-object suppression strategies in Section~\ref{sec:AEusingGT}. If we suppress non-object regions, we will affect a large fraction of search space. After we reduce curriculum factor to zero at the end of training, the network will need to re-score most of the candidates in search space. In contrast, when we only excite object regions, the network will only need to keep track of much fewer positive examples. Therefore, the model can more easily handle such a change and yield better results, as shown and compared in Table~\ref{tab:table3}.

\begin{table}
\begin{center}
\caption{The comparison results of YOLOv3+ with state-of-the-art detectors on MSCOCO2017 test dev-set.
The results for the other detectors were adapted from~\cite{yolo3,focal}.Our proposed YOLOv2+ achieved better accuracies in all terms of APs compared to the previous state-of-the-art detection results.}
%\vspace{-1mm}
\resizebox{\linewidth}{!}{
\begin{tabular}{|l|l|c c c|}
\hline
Method                                & data        & $AP$     &  $AP_{50}$   & $AP_{75}$    \\ \hline
Faster RCNN+++~\cite{resnet}        & train       & 34.9     &  55.7        &  37.4        \\ \hline
Faster RCNN w FPN~\cite{FPN}        & train       & 36.2     &  59.1        &  39.0        \\ \hline
RetinaNet\hfill(800)~\cite{focal}         & trainval35k & \textbf{40.8} & \textbf{61.1} & \textbf{44.1}   \\ \hline
YOLOv3\hfill(608)~\cite{yolo3}            & trainval35k & 33.0          & 57.9 & 34.4                     \\ \hline
YOLOv3+\hfill(608)                        & trainval35k & 35.2 & 58.4 &  38.4                                \\ \hline
\end{tabular}
}
\label{tab:table9}
\end{center}
\end{table}

%\vspace{2mm}
\noindent\textbf{What happens during back-propagation?\newline} 
Our Assisted Excitation module has an effect on back-propagation. Since AE amplifies certain activations, the effect of the receptive field gets amplified as well. Therefore, Positive examples and mislocalized examples will have a higher effect on training~(in contrast to easy negative examples that will have lower effect). This is similar to the idea behind Focal Loss. The authors show that increasing focus on positive and hard negative examples improves accuracy.

\vspace{3mm}
\noindent\textbf{Curriculum learning\newline}
Our technique is similar to curriculum learning because we start from an easier task and gradually move toward more complex tasks. However, there is a subtle difference here. Curriculum learning moves from easy to difficult by introducing increasingly difficult examples. In contrast, we move from easy to difficult by first injecting ground-truth information to the model and gradually removing this information. In other words, our tasks are easier in the initial stages not because the examples are easier, but because we help boost the correct answer. This version of curriculum learning has room to be investigated in further applications.

\vspace{3mm}
\noindent\textbf{Applicability\newline}
Our technique is applicable not only to other single-stage detectors, but also to two-stage detectors. Moreover, the AE module can be integrated in different CNN architectures for different computer vision problems, e.g., image classification(Fine-grained), segmentation, and synthesis.
\newpage
{\small
\bibliographystyle{ieee}
\bibliography{egbib}
}
\end{document}